\pdfoutput=1

\documentclass[11pt]{article}

\usepackage[]{acl}

\usepackage{times}
\usepackage{latexsym}
\usepackage{amsmath}
\usepackage{stfloats}

\usepackage[T1]{fontenc}

\usepackage[utf8]{inputenc}

\usepackage{microtype}

\usepackage{array,multirow,graphicx}
\usepackage{tabularx,colortbl}
\usepackage{background}

\SetBgContents{Accepted for publication at the Fourth Workshop on Perspectivist Approaches to NLP 2025}
\SetBgScale{1}
\SetBgAngle{0}
\SetBgPosition{current page.north}
\SetBgVshift{-1cm}

\title{Consistency is Key: Disentangling Label Variation in \\ Natural Language Processing with Intra-Annotator Agreement}

\author{Gavin Abercrombie$^{1}$ \and  Tanvi Dinkar$^{1,}$ \and Amanda Cercas Curry$^{2}$  \\
        \and \textbf{Verena Rieser}$^{1}$\thanks{\ \ Now at Google DeepMind} \and \textbf{Dirk Hovy}$^{3}$\\
        $^{1}$Heriot-Watt University  $^{2}$CENTAI Institute $^{3}$Bocconi University \\
        \texttt{g.abercrombie@hw.ac.uk}
        }

\begin{document}
\maketitle
\begin{abstract}
 
We commonly use agreement measures to assess the utility of judgements made by human annotators in Natural Language Processing (NLP) tasks.
While \emph{inter}-annotator agreement is frequently 
used as an indication of label \emph{reliability} by measuring consistency between annotators,  we argue for the additional use of \emph{intra}-annotator agreement to measure label \emph{stability} (and annotator \emph{consistency}) over time.
However, in a systematic review, we find that 
the latter is rarely reported in this field. 
Calculating these measures can act as important quality control and could provide insights into \emph{why} annotators disagree.
We conduct
exploratory annotation experiments to investigate the relationships between these measures and perceptions of subjectivity and ambiguity in text items, finding that annotators provide inconsistent responses around $25\%$ of the time across four different NLP tasks.
\end{abstract}

\section{Introduction}

Agreement measures are commonly used to assess the utility of judgements made by human annotators for Natural Language Processing (NLP) tasks.
Indeed, the reporting of \emph{inter}-annotator agreement (or inter-rater reliability)
has long been the standard to indicate dataset quality~\cite{carletta1996assessing} and frequently serves as an upper bound for model performance on a task \cite{boguslav2017inter}.

While inter-annotator agreement is 
frequently used in NLP to determine the \emph{reliability} of labels or the processes used to produce them~\citep{artstein-2017-inter}, \emph{intra}-annotator agreement is rarely, if ever, reported. However, we can use it  to measure the temporal \emph{consistency} of the annotators who chose the labels and, hence, the \emph{stability} of the labels and data that they generate.\footnote{We apply the term \emph{consistency} to annotator behaviour and \emph{stability} to labels and datasets.} 
Consistency and label stability are important because, without them, annotation schemes are unlikely to be repeatable or reproducible~\citep{teufel-etal-1999-annotation}.\footnote{Although there may be situations in which annotation consistency is not expected, such as longitudinal studies of attitudinal change.}

Such measures of intra-rater agreement are frequently reported in areas of medicine such as physiotherapy \citep[e.g.][]{bennell-etal-1998-intra,meseguer-etal-2018-inter}, and speech pathology \citep[e.g.][]{capilouto-etal-2005-ciu,rose-douglas-2003-limb}.
Intra-rater measures are also reported in other fields as diverse as economics \citep{hodgson_2008}, software engineering \citep{GRIMSTAD20071770}, and psychology~\citep{ashton-2000-review}.

However, reporting intra-annotator agreement is so far extremely uncommon in NLP, as we show in a systematic review in Section \ref{sec:review}.

\paragraph{Disagreement and label variation in NLP}
In addition, we argue that the use of inter- and intra-annotator agreement allows us to distinguish and measure different sources of observed label variation \cite{rottger-etal-2022-two,plank-2022-problem}. This is important 
as NLP researchers have increasingly recognised that, for many tasks, different points of view may be equally valid \citep{,aroyo-welty-2015-truth,basile-etal-2021-toward,plank-2022-problem,rottger-etal-2022-two}, and that their aggregation can erase minority perspectives \citep{basile-etal-2021-toward,blodgett-2021-socio}.\looseness=-1 

One of the main challenges in implementing this new paradigm is the interpretation of disagreement. Disagreement between annotators may be due to two sources: 
1) genuine differences in their subjective beliefs/perspectives, which can be desirable under this paradigm, or 
2) task difficulty, ambiguity, or annotator error, all of which are undesirable.
While agreement measures \emph{between} annotators can give us an idea of task \textbf{subjectivity}, they provide little insight as to its \textbf{difficulty}, \textbf{ambiguity,} or the quality and attentiveness of the annotators themselves~\citep{rottger-etal-2022-two}.

In the following, we propose the use of {\em intra}-annotator agreement as a measure of subjectivity.

\paragraph{The reliability-stability agreement matrix}

What then, does it mean when individual annotators' interpretations are not stable, i.e., internally inconsistent?
In addition to providing an additional layer of quality control, we suggest that measurement of label stability can help to interpret potential causes of \emph{inter}-annotator disagreement.
To this end, we propose the reliability-stability matrix, a framework for mapping and interpreting the relationship between inter- and intra-annotator agreement in labelled datasets (Table \ref{tab:matrix}).

\begin{table}[ht!]
    \centering
    \small
    \begin{tabular}{cc|c|c}
                             &  & \multicolumn{2}{c}{\textbf{Reliability}}  \\
                             &  & \multicolumn{2}{c}{(between annotators)} \\
                             &  & Low \emph{inter}  & High \emph{inter} \\

        \hline
        \multirow{6}{*}{\rotatebox[origin=c]{90}{\parbox{15mm}{\textbf{Stability} (temporal within annotator)}}}
        & \multirow{2}{*}{High}  &  Variable & \cellcolor[gray]{.9} Straight- \\
        
         & \multirow{2}{*}{\emph{intra}} &  perspectives/ & \cellcolor[gray]{.9} forward/ \\ 

         &  & High subjectivity & \cellcolor[gray]{.9} Good quality \\

         \cline{2-4}
         & \multirow{2}{*}{Low}   & \cellcolor[gray]{.9} Ambiguous & \cellcolor[gray]{.7} Systematic  \\
         & \multirow{2}{*}{\emph{intra}} & \cellcolor[gray]{.9} or difficult/ & \cellcolor[gray]{.7} errors/   \\

         &  & \cellcolor[gray]{.9} Poor quality & \cellcolor[gray]{.7} Value changes
    \end{tabular}
    \normalsize
    \caption{The reliability-stability matrix for \emph{inter-} and \emph{intra-} annotator agreement.}
    \label{tab:matrix}
\end{table}

Under this framework, \emph{inter}-annotator agreement and  \emph{intra}-annotator agreement, taken together, indicate the task's ambiguity or complexity and its subjectivity level.
\emph{Inter}-annotator agreement measures reliability, while \emph{intra}-annotator agreement measures stability. 
The resulting axes form a confusion matrix that describes four cases.

If both measures are high, we assume the task is unambiguous and simple, and the annotator group relatively homogonous. 
Presumably, the quality of the guidelines and textual data is also good \cite{ide2017handbook}. 
In this scenario, the task or item should be relatively straightforward.

Where both agreement measures are low, we are likely to be faced with a highly ambiguous or difficult task or item--perhaps with multiple equally valid responses--or the annotation quality is poor.

If reliability is low, but consistency is high, the labels likely reflect the annotators' varied but potentially equally valid subjective perspectives.

We do not foresee many situations where reliability is high
yet stability/consistency is low. Any agreement between inconsistent annotators would presumably be purely by chance or mass random spamming, i.e., systematic errors.
Exceptions could include population-level value shifts over longer time intervals arising from awareness-raising events such as the \#MeToo~\citep{szekeres-etal-2020-views} and \#BLM~\citep{sawyer-gampa-2018-implicit} movements.\looseness=-1 

Our framework can be applied at the dataset- or item-level by computing 
any standard agreement metrics.\looseness=-1
We  illustrate this in exploratory annotation experiments described in Section \ref{sec:experiments}.

\noindent\paragraph{Our contributions}
1) We conduct a systematic review, finding that \textbf{a tiny fraction of NLP publications report intra-annotator agreement};
(2) we suggest addition of intra-annotator agreement as a standard measure, and show how \textbf{measuring annotator stability could complement existing reliability measures to distinguish reasons for label variation};
and (3) we conduct exploratory longitudinal annotation experiments across four NLP tasks, finding that \textbf{annotators provide inconsistent responses for more than {\boldmath$25\%$} of items}, calling into question the implicit assumption that differences in annotation behaviour are seen only between and not within individuals.\footnote{Data available at \url{https://github.com/HWU-NLP/consistency}.}

\section{Intra-Annotator Agreement in the NLP Community} \label{sec:review}

To get a snapshot of the extent to which intra-annotator agreement is reported in the NLP community, we conducted a systematic review of papers published in the \emph{Anthology} of the Association for Computational Linguistics (ACL).\footnote{\url{https://aclanthology.org/}}
Here, we wish to discover for which tasks and what purposes NLP researchers collect and report on repeat annotations and evidence for how and when repeat items should be presented to annotators.
Full details of the review methodology are available in Appendix~\ref{app:slr}.

\begin{table*}[hb!]
    \centering
    \begin{tabular}{l|lll}
         & \textbf{Task} & \textbf{Dataset} & \textbf{Labels} \\
         \hline 
        \multirow{2}{*}{\textbf{Social}} & Offensive language detection & \citet{leonardelli-etal-2021-agreeing} & \emph{Offensive}/\emph{not offensive} \\ 
         & Sentiment analysis & \citet{kenyon-dean-etal-2018-sentiment} & \emph{Positive}/\emph{negative}/\emph{objective} \\
        \hline 
          & Natural language inference/  & \multirow{2}{*}{\citet{williams-etal-2018-broad}}  & \emph{Entailment}/\emph{contradiction}/  \\
         \textbf{Linguistic} & textual entailment &  & \emph{neutral} \\
          & Anaphora resolution & \citet{poesio-etal-2019-crowdsourced} & \emph{Referring}/\emph{non-referring}
    \end{tabular}
    \normalsize
    \caption{Datasets used in the annotation experiments.}
    \label{tab:datasets}
\end{table*}

\paragraph{To what extent and why is intra-annotator agreement reported in NLP?}
When we conducted our study, the search and filtering process returned only 56 relevant publications out of more than 80,000 papers listed in the Anthology.
In other words, a tiny fraction (less than 0.07\%) of computational linguistics and NLP publications in the repository report measurement of intra-annotator agreement.\footnote{We acknowledge that intra-annotator agreement is irrelevant to many papers, but highlight that the number of publications which report it is nevertheless extremely low.}

The only area of NLP in which intra-annotator agreement is somewhat regularly reported is machine translation (MT), which accounts for more than half of the included publications.
Most of these were agreement measures on human evaluation of translation quality, with one on word alignment annotation for MT~\citep{li-etal-2010-enriching}.
Several other publications on evaluating natural language generation also report measurement on human evaluation tasks~\citep[e.g.][]{belz-kow-2011-discrete,belz-etal-2016-effect,belz-etal-2018-spatialvoc2k,jovanovic-etal-2005-corpus}.
Other included fields are semantics~\citep[e.g.][]{cao-etal-2022-theory,hengchen-tahmasebi-2021-supersim}, syntax~\citep[e.g.][]{baldridge-palmer-2009-well,lameris-stymne-2021-whits}, affective computing (including sentiment analysis~\citep{kiritchenko-mohammad-2017-best} and emotion detection~\citep{vaassen-daelemans-2011-automatic}), and automatic text grading~\citep{cleuren-etal-2008-childrens,downey-etal-2011-performance}.
There is also one paper on abusive language detection~\citep{cercas-curry-etal-2021-convabuse}.\footnote{We provide a full list of included papers in Appendix \ref{app:included_papers}.}\looseness=-1

Where the authors motivate the collection of repeat annotations, they usually mention quality control or annotator consistency. Notably, no papers mention the possibility that intra-annotator inconsistency could be valid or informative beyond these factors, as we propose.

\paragraph{Best practice for measuring intra-annotator agreement: how long should the label-relabel interval be?}

When designing annotation tasks (such as ours in Section \ref{sec:experiments}), it would be helpful to know when to present repeated items, thus avoiding  annotators labelling from memory, which may not be an actual test of their consistency.

Over a quarter of the papers (15/56) do not provide enough information to determine the interval between initial and repeat annotations.
In most other cases, either it can be inferred, or the authors explicitly state that re-annotations are conducted in the same session as the original annotation.
Those that report more extended time before re-annotation leave intervals varying from a few minutes \citep{kiritchenko-mohammad-2017-best} to a year \citep{cleuren-etal-2008-childrens,hamon-2010-judge}.

Two papers do specifically investigate the effects of time on annotator consistency.
\citet{li-etal-2010-enriching} experimented with intervals of one week, two weeks, and one month, comparing intra-annotator agreement for these and finding that consistency on their word alignment annotation degraded steadily over time.
\citet{kiritchenko-mohammad-2017-best} performed a similar study, comparing intra-annotator agreement on ratings (on a scale) that were conducted with intervals from a few minutes to a few
days between the initial and repeat judgements.
They too found that inconsistencies increased as a function of increase in interval.

\section{Exploratory annotation experiments} \label{sec:experiments} 
We conduct an exploratory annotation experiment to investigate the relationships between agreement measures and the possible reasons for disagreements and inconsistencies.
We also investigate whether, as is commonly believed, specific task types are generally more subjective than others.

\vspace{0.35cm}

\noindent\textbf{Hypotheses}

\noindent At the individual annotation item level, for a given task and dataset:

\begin{itemize}

\item[H1.1] \emph{Subjective} annotation items have lower \emph{inter}-annotator agreement than \emph{straightforward} items, but higher \emph{intra-}annotator agreement than \emph{ambiguous} items.

\item[H1.2] \emph{Ambiguous} annotation items have lower \emph{inter}- and \emph{intra-}annotator agreement than both \emph{straightforward} and \emph{ambiguous} items. 

\end{itemize}

\noindent At the dataset/task level:

\begin{itemize}
    \item[H2] \emph{Social} tasks---such as offensive language detection and sentiment analysis---are more \emph{subjective} than \emph{linguistic} tasks, like textual entailment or anaphora resolution.
    That is, stability is higher for social tasks than linguistic tasks.
\end{itemize}

\paragraph{Data}

We use subsets of four English language datasets, see Table \ref{tab:datasets}: two social tasks that are commonly assumed to be subjective, and two linguistic tasks, thought of as objective \citep{basile-etal-2021-need}.
These were selected because they (1) have limited label sets (of two or three classes), allowing for comparison across tasks; and (2) have been published with non-aggregated (i.e.\ annotator specific) labels, allowing us to include items with known inter-annotator disagreement in our subsamples. 
From each dataset, we selected 50 items with high disagreement in the original label sets for re-annotation.\looseness=-1

\paragraph{Methodology}
We recruited 
crowdworkers from Prolific\footnote{\url{https://www.prolific.co/}} to annotate a subset of 
fifty items 
from each of the tasks/datasets. 
As much of the text data is primarily sourced from the United States of America and, in some cases,\footnote{Particluarly in the offensive language dataset.} concerns American news stories such as the controversy surrounding the killing of George Floyd,\footnote{\emph{The Guardian} April 20 2021 \citep{mcgreal-2021-derek}.} we recruited only annotators located in the US.
To obtain high quality annotations, we prescreened participants to ensure that (1) their first language was English, and that (2) they had a $100\%$ approval rate on Prolific.

Based on the evidence of our review~\citep{li-etal-2010-enriching,kiritchenko-mohammad-2017-best},
and of more recent work by \citet{abercrombie-etal-2023-temporal},
we left an interval of two weeks before we recall the annotators to collect a second round of annotations in order to measure their consistency.
Of 30 annotators that began the first task, 16 completed both rounds of all four tasks, and we base our results on the labels they provided.
All annotators were L1 English speakers; nine were male and eight female; 11 identified as `White', four as `Black', one `Asian', and one `Mixed';
and ages ranged from 20 to 67; ($\mu=43.9; s=14.0$).
Annotators were provided with the original instructions pertaining to each task.

We then recruited a second set of expert annotators to annotate the examples that demonstrate internal and or external disagreement with rationalisations for these disagreements, using the labels  
\textit{ambiguous}, 
\textit{subjective}, or \emph{straightforward}.

\section{Results}

We report agreement for each task, 
and examine differences between the groups of items labelled as \emph{subjective}, \emph{ambiguous}, and \emph{straightforward}.\looseness=-1

\paragraph{Overall agreement}

As \emph{intra}-annotator agreement is typically assumed to be $100\%$ (i.e. by omitting to consider it~\citep{abercrombie-etal-2023-temporal}), we measure and raw report percentage agreement as a primary metric to examine whether this holds.
For \emph{inter}-annotator agreement, we calculate these pairwise across  annotators and report the means.
For completeness, we also report Cohen's kappa scores in \autoref{app:cohens}.

\begin{table}[ht!]
    \centering
    \begin{tabular}{l|rr|rr}
                & \multicolumn{2}{c|}{\textbf{Reliability}} & \multicolumn{2}{c}{\textbf{Stability}}  \\
                & \multicolumn{2}{c|}{(\emph{Inter-})} & \multicolumn{2}{c}{(\emph{Intra-})} \\
                & \multicolumn{2}{c}{$\%$} & \multicolumn{2}{c||}{$\%$} \\
                & $\mu$ & $\sigma$ & $\mu$ & $\sigma$ \\
        \hline 
        Offence    & 68.3 & 15.4  & 74.4 & 15.0      \\
        Sentiment  & 63.6 & 21.7  & 69.2 & 19.5  \\
        Entailment & 58.6 & 21.4  & 72.6 & 15.1  \\
        Anaphora   & 76.2 & 14.3  & 80.5 & 13.0  \\
        \hline 
        Overall  & 66.7 & 19.6  & 74.2 & 16.3 
    \end{tabular}
    \caption{Pairwise reliability and stability of the collected labels measured with mean ($\mu$) and standard deviations ($\sigma$) across items for raw percentage inter- and intra-annotator agreement scores.}
    \label{tab:prolific_agreement}
\end{table}

Agreement scores are presented in Table \ref{tab:prolific_agreement}.
As expected, agreement is higher for stability than reliability for all tasks, although considerably lower than perfect agreement---just $74.2\%$ overall, and no higher than $80.5\%$ for any task.
Individual annotators all have very similar levels of stability: $\mu=74.2\%; \sigma=4.3\%; max=81.5\%; min=67.5\%$.
These results are also remarkably similar to those of \citet{abercrombie-etal-2023-temporal}, who reported mean intra-annotator agreement of $74.5\%$ on a hate speech identification task conducted over a comparable time frame and on the same recruitment and annotation platforms. 

\paragraph{Agreement by task}

\begin{figure}[ht!]
    \centering
    \includegraphics[width=\linewidth]{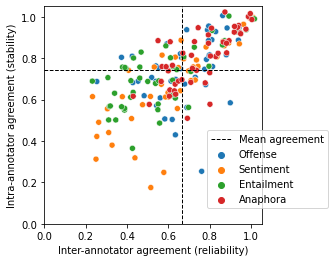}
    \caption{By task raw percentage agreement on individual items for reliability (pairwise) and stability.}
    \label{fig:tasks}
\end{figure}

The distribution of annotation items on the reliability-stability matrix is shown in \autoref{fig:tasks}.
A multivariate Kruskal-Wallis test indicates statistically significant differences between tasks for both variables: 
for inter-annotator agreement, $H-statistic: 12.42, p-value: 0.01$; and for intra-annotator agreement, $H-statistic: 10.76, p-value: 0.01$.\footnote{Post-hoc pairwise Dunn's tests with Bonferroni correction reveal that only \emph{sentiment-anaphora} and \emph{entailment	anaphora} have significantly different distributions for reliability, and only \emph{sentiment-anaphora} for stability.}

However, these differences do not confirm the view that social tasks are more subjective than linguistic tasks (H2). 
Rather, the \emph{offense} and \emph{anaphora} tasks obtain higher agreement (both \emph{inter} and \emph{intra}) than the \emph{sentiment} and \emph{entailment tasks}, suggesting that, for the particular items in these data samples, the former are simply easier to agree and be consistent on than the latter.

\begin{table}[ht!]
    \centering
    \small
    \begin{tabular}{l|cccc}
          & Bottom- & Top- & Top- & Bottom- \\
          & left & left & right & right \\
          & (\emph{Amb.}) & (\emph{Subj.}) & (\emph{Straight.}) & (\emph{Errors}) \\
         \hline 
         Offense & 30.0 & 18.0 & 38.0 & 14.0 \\
Sentiment & 46.0 & 10.0 & 38.0 & 6.0 \\
Entailment & 48.0 & 20.0 & 28.0 & 4.0 \\
Anaphora & 22.0 & 10.0 & 56.0 & 12.0 \\
\hline 
Overall & 36.5 & 14.5 & 40.0 & 9.0 
    \end{tabular}
    \caption{Percentage of annotation items in each quadrant of the plot in \autoref{fig:tasks}.}
    \label{tab:tasks}
\end{table}

As \autoref{fig:tasks} and \autoref{tab:tasks} show, 
while the annotation items are predominantly distributed across the bottom-left and top-right quadrants, \emph{sentiment} and \emph{entailment} are skewed to the bottom left, indicating greater ambiguity, and offensive language and entailment tend towards the top-right (\emph{subjectivity}).
With $68\%$ of items on the left-hand side, \emph{entailment} is the least, and \emph{anaphora}, with $56\%$ in the top-right, the most \emph{straightforward} task.

Anaphora resolution seems to be the most \emph{straightforward} task, with most items in the upper-right quadrant, while sentiment analysis and entailment are the most ambiguous/difficult, both having almost $50\%$ of examples fall in the bottom left quadrant.
As expected, the lowest number of items fall in the bottom right section of the plot.

\paragraph{Rationalisation} 
In an attempt to validate the reliability-stability matrix and to test H1.1 and H1.2, rationalisation labels were applied by two postdoctoral researchers with backgrounds in NLP and computational linguistics.
They were asked to read the annotation instructions and items and provide each example with a label: \emph{subjective}, \emph{ambiguous}, or \emph{straightforward}. 
Disagreements were resolved by discussion between these and a third author.
Inter-annotator agreement (before resolution) is shown in \autoref{tab:expertagree}, indicating that this in itself was a very difficult task to reach agreement on.\looseness=-1

\begin{table}[ht!]
    \centering
    \begin{tabular}{rrrr}
        Offence & Sentiment & Entailment & Anaphora \\
        \hline
        0.26 & 0.11 & 0.47 & 0.02 
    \end{tabular}
    \caption{Inter-annotator agreement on the rationalisation labelling task, measured with Cohen's \emph{kappa}.}
    \label{tab:expertagree}
\end{table}

To quantitatively examine the relationship between the perceived reason for agreement/disagreement 
and the reliability and stability measurements,
we 
applied a multivariate Kruskal-Wallis test
to the independent categorical variable \emph{rationale} (\emph{straightforward}, \emph{subjective}, and \emph{ambiguous}) and the two dependent continuous variables \emph{inter-} and \emph{intra-annotator agreement}.

The test showed that there is only a very small and non-significant difference in the dependent vectors between the different groups, with an $H$-statistic of $2.734, p=0.26$, indicating that the assigned rationale labels do not explain the inter- and intra annotator agreement rates.

\section{Discussion and conclusion}

We have examined the role and use of intra-annotator agreement measures in NLP research.
Calculation of such measures can act as an important quality control and could potentially provide insights into the reasons for disagreements between annotators.
However, in a systematic review, we found that they are rarely reported in this field.

We have proposed a framework for the interpretation of inter- and intra-annotator agreement, the \emph{reliabilty-stability agreement matrix}.
Exploratory annotation experiments failed to validate our theory that this framework can be used to tease apart subjectivity and ambiguity, and it proved to be very hard to recognise or agree on these, even for trained annotators. 
However, we have shown how comparing both \emph{inter-} and \emph{intra-} annotator agreement enables quantification of the difficulty of particular tasks and/or annotation items.
Strikingly, we found that, across four different tasks, crowdsourced annotators were consistently inconsistent, calling into question the implicit assumption that labels provided by individual annotators are stable, and reinforcing the need to collect within-annotator labels for NLP tasks, including those typically considered to be `objective'.

\section*{Limitations}

We acknowledge that the scope of our exploratory experiments is quite small at 50 items per task and 16 annotators, and that larger studies may produce different results.
While we took some measures to ensure the quality of recruited annotators (\autoref{sec:experiments}), there are known issues with crowdworker quality for annotation~\citep[e.g.][]{hovy-etal-2013-learning,weber-genzel-etal-2024-varierr}, and some annotator inconsistency may due to inattention---another factor that should be considered and further reason to measure and report intra-annotator agreement.

\section*{Ethical considerations}

Because we recruit humans to work on data labelling, we obtained approval to undertake this study from the Institutional Review Board (IRB)
of the School of Mathematics \& Computer Science at Heriot-Watt University, reference 2023-4926-7368.
Additionally, we took the following measures:

\paragraph{Compensation} We paid the annotators above the Living Wage in our jurisdiction (higher than the legal minimum wage, as recommended (as a minimum) by~\citet{shmueli-etal-2021-beyond}.

\paragraph{Welfare}
As some of the data to be labelled included offensive language, we:
\vspace{-0.2cm}
\begin{itemize}
    \item avoided recruiting members of vulnerable groups by restricting annotators to those aged over 18, provided them with comprehensive warnings prior to consenting to participate, and asked them to self-declare that they would not be adversely affected by participating;
    \item allowed annotators to leave the study at any time and informed them that they would be  paid for their time regardless;
    \item kept the annotation task short to avoid lengthy exposure to material which may exceed `\emph{minimal risk}' \citep{shmueli-etal-2021-beyond}.
\end{itemize}

\paragraph{Privacy} All personal data of recruited annotators was collected anonymously. 

\section*{Acknowledgements}
We would like to thank the reviewers for their insightful comments which we have tried to incorporate into this version of the paper.

Gavin Abercrombie and Tanvi Dinkar were supported by the EPSRC project ‘Equally Safe Online’ (EP/W025493/1).
Dirk Hovy was supported by the European Research Council (ERC) under the European Union's Horizon 2020 research and innovation program (grant agreement No. 949944, INTEGRATOR).
He is a member of the MilaNLP group and the
Data and Marketing Insights Unit of the Bocconi
Institute for Data Science and Analysis (BIDSA).

\bibliography{anthology,custom}
\bibliographystyle{acl_natbib}

\appendix

\section{Systematic review methodology}
\label{app:slr}

For this review, we followed the established systematic review guidelines of the PRISMA (Preferred Reporting Items for Systematic Reviews and Meta-Analyses) statement \citep{moher2009preferred}, as recommended by \citet{van-miltenburg-etal-2021-preregistering}:

\begin{enumerate}
    \item Develop search query terms
    \item Conduct search
    \item Apply inclusion/exclusion criteria
    \item Code included publications
    \item Measure inter- and intra-annotator agreement (re-code subset of publications)
    \item Synthesise results
\end{enumerate}

The review covers all results retrieved from the Anthology's search facility. 
The searches were conducted on September 14 2022.
Following retrieval of the resulting publications, we applied the inclusion/exclusion criteria shown in Table \ref{tab:inclusion_exclusion}.

\begin{table}[ht!]
    \centering
    \small
    \begin{tabular}{p{3.5cm} p{3.5cm}}
        \textbf{Include} & \textbf{Exclude} \\
        \hline
        Human annotation studies & No human annotation study is conducted (e.g. surveys/reviews of other work) \\
        Repeated annotations are collected & Repeated annotations are not collected \\
        Intra-annotator measurement is reported & Intra-annotator measurement not reported \\
        Measurement conducted on manual labels applied by human annotators & Labelling is performed automatically \\
        `Intra-' refers to repeat annotations of the same items by the same annotator & Term `intra-' is used, but refers to agreement measurements between different items and/or annotators \\
        Publication is a full paper & Posters, proceedings, proposals, technical system descriptions etc.
    \end{tabular}
    \normalsize
    \caption{Criteria for in/exclusion in/from the review.}
    \label{tab:inclusion_exclusion}
\end{table}

The searches returned 138 publications. 
After removing duplicates, and applying the inclusion criteria we were left with 56 relevant publications in the Anthology.

\clearpage

\section{Included papers} \label{app:included_papers}

A list of included publications from the ACL Anthology that report intra-annotator agreement is presented in Table \ref{tab:papers}. 

\begin{table*}[ht!]
    \centering
    \small
    \begin{tabular}{ll|ll}
        \textbf{Publication} & \textbf{NLP sub-field} & \textbf{Publication} & \textbf{NLP sub-field} \\
        \hline 
        \citet{akhbardeh-etal-2021-findings} & Machine Translation & \citet{graham-etal-2013-continuous} & Machine Translation  \\
        \citet{alosaimy-atwell-2018-web} & Syntax & \citet{grundkiewicz-etal-2015-human} & Syntax \\
        \citet{baldridge-palmer-2009-well} & Machine Translation & \citet{hamon-2010-judge} & Machine Translation  \\
        \citet{belz-kow-2011-discrete} & NLG & \citet{he-etal-2010-improving} & Machine Translation \\
        \citet{belz-etal-2016-effect} & NLG & \citet{hengchen-tahmasebi-2021-supersim} & Semantics \\
        \citet{belz-etal-2018-spatialvoc2k} & NLG & \citet{herbelot-copestake-2010-annotating} & Semantics \\
        \citet{bentivogli-etal-2011-getting} & Machine Translation & \citet{hochberg-etal-2014-decision} & Cognitive psychology \\
        \citet{berka-etal-2011-quiz} & Machine Translation & \citet{hochberg-etal-2014-towards} & Cognitive psychology \\
        \citet{bojar-etal-2013-findings} & Machine Translation & \citet{jovanovic-etal-2005-corpus} & NLG \\
        \citet{bojar-etal-2014-findings} & Machine Translation & \citet{kiritchenko-mohammad-2017-best} & Affective computing \\ 
        \citet{bojar-etal-2015-findings} & Machine Translation & \citet{kreutzer-etal-2020-correct} & Machine Translation \\
        \citet{bojar-etal-2016-findings} & Machine Translation & \citet{kreutzer-etal-2018-reliability} & Machine Translation \\
        \citet{bojar-etal-2017-findings} & Machine Translation & \citet{kruijff-korbayova-etal-2006-annotation} & Machine Translationn \\
        \citet{bojar-etal-2018-findings} & Machine Translation & \citet{lameris-stymne-2021-whits} & Syntax \\
        \citet{bouamor-etal-2014-human} & Machine Translation & \citet{laubli-etal-2013-statistical} & Machine Translation \\
        \citet{callison-burch-etal-2007-meta} & Machine Translation & \citet{li-etal-2010-enriching} & Machine Translation \\
        \citet{callison-burch-etal-2010-findings} & Machine Translation & \citet{long-etal-2020-shallow} & Semantics \\
        \citet{callison-burch-etal-2011-findings} & Machine Translation & \citet{mccoy-etal-2012-annotation} & Cognitive psychology \\
        \citet{callison-burch-etal-2012-findings} & Machine Translation & \citet{ruiter-etal-2022-exploiting} & NLG \\
        \citet{callison-burch-etal-2009-findings} & Machine Translation & \citet{sanchez-cartagena-etal-2012-source} & Machine Translation \\
        \citet{callison-burch-etal-2008-meta} & Machine Translation & \citet{schulz-etal-2019-analysis} & Semantics \\
        \citet{cao-etal-2022-theory} & Semantics & \citet{vaassen-daelemans-2011-automatic} & Affective computing \\
        \citet{cercas-curry-etal-2021-convabuse} & Abuse detection & \citet{vela-van-genabith-2015-assessing-wmt2013} & Machine Translation \\
        \citet{cleuren-etal-2008-childrens} & Automatic text grading & \citet{walsh-etal-2020-annotating} & Syntax \\
        \citet{dsouza-etal-2021-semeval} & Sematics & \citet{wang-sennrich-2020-exposure} & Machine Translation \\
        \citet{deshpande-etal-2022-stereokg} & Semantics & \citet{wang-etal-2021-exploring} & Machine Translation  \\
        \citet{downey-etal-2011-performance} & Automatic text grading & \citet{wang-etal-2014-query} & NLG  \\
        \citet{friedrich-palmer-2014-situation} & Semantics & \citet{zeyrek-etal-2018-multilingual} & Semantics \\
        
    \end{tabular}
    \normalsize
    \caption{Publications in the ACL Anthology in which intra-annotator agreement is reported.}
    \label{tab:papers}
\end{table*}

\section{Cohen's kappa scores} \label{app:cohens}

\begin{table}[h!]
    \centering
    \begin{tabular}{l|rr|rr}
                & \multicolumn{2}{c|}{\textbf{Reliability}} & \multicolumn{2}{c}{\textbf{Stability}}  \\
                & \multicolumn{2}{c|}{(\emph{Inter-})} & \multicolumn{2}{c}{(\emph{Intra-})} \\
                & \multicolumn{2}{c|}{$\kappa$} & \multicolumn{2}{c}{$\kappa$} \\
                & $\mu$ & $\sigma$ & $\mu$ & $\sigma$ \\
        \hline 
        Offence    & 0.05 & 0.28 &  0.27 & 0.28    \\
        Sentiment  & 0.02 & 0.25 & 0.17 & 0.29  \\
        Entailment & 0.02 & 0.25 & 0.28 & 0.28  \\
        Anaphora   & 0.07 & 0.31 & 0.22 & 0.35   \\
        \hline 
        Overall  & 0.04 & 0.28 &  0.23 & 0.30
    \end{tabular}
    \caption{Pairwise reliability and stability of the collected labels measured with mean ($\mu$) and standard deviations ($\sigma$) across items for inter- and intra-annotator agreement scores measured with Cohen's kappa ($\kappa$).}
    \label{tab:prolific_agreement}
\end{table}

\end{document}